\documentclass[sigconf,nonacm]{acmart}
\renewcommand\footnotetextcopyrightpermission[1]{}
\AtBeginDocument{%
  }

\AtEndPreamble{
    \usepackage[capitalize]{cleveref}
    \crefname{section}{Sec.}{Secs.}
    \Crefname{section}{Section}{Sections}
    \Crefname{table}{Table}{Tables}
    \crefname{table}{Tab.}{Tabs.}
}

\usepackage{xspace}
\makeatletter
\DeclareRobustCommand\onedot{\futurelet\@let@token\@onedot}
\def\@onedot{\ifx\@let@token.\else.\null\fi\xspace}

\def\ie{\emph{i.e}\onedot}

\def\etal{\emph{et al}\onedot}
\makeatother

\RequirePackage[shortlabels,inline]{enumitem}
\setlist[itemize]{noitemsep,leftmargin=*,topsep=0em}
\setlist[enumerate]{noitemsep,leftmargin=*,topsep=0em}

\usepackage{multirow}
\usepackage{pifont}
\usepackage{amsmath}
\DeclareMathOperator{\atanTwo}{atan2}
\usepackage{bibunits}
\defaultbibliographystyle{ACM-Reference-Format}
\usepackage{caption}

\begin{document}

\title{Deep Light Pollution Removal in Night Cityscape Photographs}

\author{Hao Wang}
\affiliation{
  \institution{Shandong University}
  \city{Qingdao}
  \country{China}
}
\email{haowang_1999@mail.sdu.edu.cn}

\author{Xiaolin Wu}
\affiliation{
  \institution{Southwest Jiaotong University}
  \city{Chengdu}
  \country{China}
}
\email{xlw@swjtu.edu.cn}

\author{Xi Zhang}
\affiliation{
  \institution{Tongji University}
  \city{Shanghai}
  \country{China}
}
\email{xzhang9308@gmail.com}

\author{Baoqing Sun}
\affiliation{
  \institution{Shandong University}
  \city{Qingdao}
  \country{China}
}
\email{baoqing.sun@sdu.edu.cn}



\begin{abstract}
Nighttime photography is severely degraded by light pollution induced by pervasive artificial lighting in urban environments. After long-range scattering and spatial diffusion, unwanted artificial light overwhelms natural night luminance, generates skyglow that washes out the view of stars and celestial objects and produces halos and glow artifacts around light sources. 
Unlike nighttime dehazing, which aims to improve detail legibility through thick air, the objective of light pollution removal is to restore the pristine night appearance by neutralizing the radiative footprint of ground lighting. In this paper we introduce a physically-based degradation model that adds to the previous ones for nighttime dehazing two critical aspects; (i) anisotropic spread of directional light sources, and (ii) skyglow caused by invisible surface lights behind skylines. In addition, we construct a training strategy that leverages large generative model and synthetic-real coupling to compensate for the scarcity of paired real data and enhance generalization. Extensive experiments demonstrate that the proposed formulation and learning framework substantially reduce light pollution artifacts and better recover authentic night imagery than prior nighttime restoration methods.
\end{abstract}



\maketitle

\begin{bibunit}
\section{Introduction}
\label{sec:intro}
\begin{figure}
    \centering
    \includegraphics[width=1.0\linewidth]{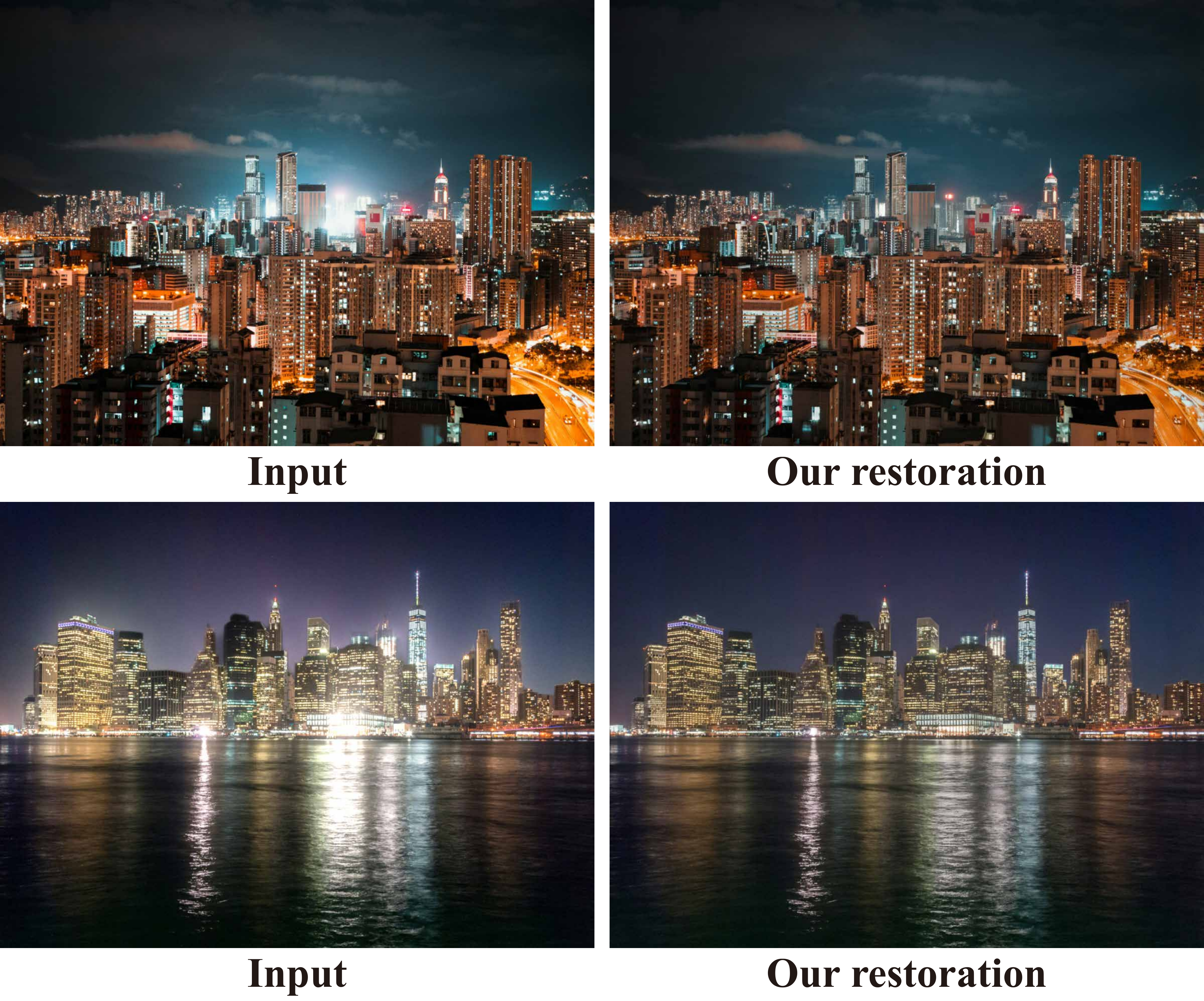}
    \vspace{-6mm}
    \caption{Examples of light-polluted images and our restoration results, showing effective light pollution removal with a natural nighttime appearance.}
    \Description{Examples of light pollution and restoration photographs.}
    \vspace{-6mm}
    \label{fig:teaser}
\end{figure}

Rapid urbanization and the widespread use of artificial lighting have made nighttime light pollution a global environmental concern. Artificial illumination, after being scattered and diffused in the atmosphere, alters both the radiance level and spectral distribution of the night environment and severely degrades the quality of nighttime photographs. Even a modest level of light pollution can ruin the artistic quality of city skylines or night skies: the long exposure required to capture dim nocturnal details also integrates unwanted radiance, yielding greyish sky backgrounds, halos around luminaires, washed-out architectural contours, and loss of midtone structures. For applications such as visual arts, HDR imaging, cultural documentation, environmental studies, and astronomy-related public outreach, it is highly desirable to restore the natural nocturnal appearance without relocating to light-free regions.

Because light pollution is radiative rather than transmissive in origin, it cannot be physically eliminated after image capture. The only remedy is algorithmic removal, which requires understanding and modeling its formation. Prior work on nighttime image processing has predominantly focused on low-light enhancement and nighttime dehazing~\cite{land1977retinex,fu2016weighted,guo2016lime,wei2018deep,zhang2019kindling,guo2020zero,xu2022snr,ma2022toward,cai2023retinexformer,fu2023learning,zhou2024glare,jiang2024lightendiffusion,ma2025incorporating,yan2025hvi,feijoo2025darkir,ancuti2016night,zhang2014nighttime,zhang2017fast,ancuti2020day,zhang2020nighttime,yan2020nighttime,liu2022nighttime,wang2022variational,jin2023enhancing,cong2024semi,lin2025nighthaze,deng2025real}. However, these tasks differ fundamentally from light pollution removal. Dehazing formulations treat the atmosphere as a passive medium that attenuates and scatters incoming light, whereas light pollution is caused by emissive and often highly directional artificial sources. Moreover, existing nighttime restoration methods rarely account for \emph{sky glow}, \ie the elevation of background sky luminance induced by urban lighting behind the skyline, which may be invisible as explicit light sources in the image but nevertheless dominates the appearance of many night cityscapes~\cite{liu2021light,narasimhan2003shedding}. Such radiance contamination is also unfavorable for downstream image representation and compression, since it alters the natural luminance structure and perceptually important content of the captured scene~\cite{lee2020training,patel2021saliency,zhang2021attention,zhang2022multi,zhang2023lvqac,cao2024learned}.

Beyond low-light enhancement and dehazing, several related directions have addressed other nighttime degradations. Flare and glare removal methods study optical reflections, halos, and veiling artifacts produced by strong light sources~\cite{talvala2007veiling,vitoria2019automatic,wu2021train,qiao2021light,dai2023nighttime,dai2023mipi,dai2024flare7k++,dai2024mipi,ghodesawar2023deflare,ma2025self}, while HDR reconstruction methods aim to recover clipped highlights and missing radiance in saturated regions~\cite{dille2024intrinsic}. These lines of work are clearly relevant, yet real nighttime cityscape photographs often exhibit several degradations simultaneously: atmospheric glow spreads around artificial lights, flare overlaps with scene structures, strong exposure imbalance suppresses shadow details, and saturation destroys textures near bright luminaires. As a result, directly applying a method specialized for only one subproblem often leads to limited correction, halo remnants, over-enhanced contrast in non-polluted regions, or unnatural nighttime appearance.

Compared with nighttime dehazing, the setting considered in this work is fundamentally different in both physics and objective. Dehazing methods address visibility degradation caused by passive media and typically seek to improve scene clarity and contrast. In contrast, our goal is to neutralize actively emitted, anisotropically propagated light and to recover an aesthetically faithful nocturnal appearance rather than merely increasing visibility. Dehazing priors and enhancement-oriented architectures, when applied directly, often under-correct sky glow, exaggerate local contrast, or leave visible contamination around luminaires.

These observations suggest that \emph{light pollution removal} should be treated as a distinct nighttime restoration problem. A preliminary step in this direction was taken by Liu and Wu~\cite{liu2021light}, who explicitly considered light pollution reduction in nighttime photography. More recent progress in nighttime flare removal and realistic nighttime restoration further highlights the diversity and difficulty of light-induced artifacts in practical scenes~\cite{dai2024flare7k++,jiang2024lightendiffusion}. Meanwhile, modern generative and large-model priors have shown strong potential for solving challenging image restoration and editing problems under complex real-world degradations~\cite{labs2025flux1kontextflowmatching,wu2025qwenimagetechnicalreport,zhang2025badiff}. However, these models are not specifically designed for the physical characteristics of nighttime light pollution and may alter scene semantics or hallucinate structures when used without task-specific constraints.

To address these limitations, we study \emph{deep light pollution removal} for nighttime cityscape photographs. Our objective is to suppress pollution caused by artificial lighting while preserving the genuine scene illumination, structure, and atmosphere of the original night view. We introduce a physically grounded image formation model that explicitly accounts for anisotropic point spread functions of directional light sources and skyline-induced sky glow, and combine physically simulated supervision with large-model priors to improve robustness and generalization on real cityscape photographs. Different from conventional low-light enhancement, we do not simply brighten dark regions; different from standalone flare removal, we do not restrict the problem to optical flare artifacts alone. Instead, we target a practical and comprehensive setting in which glare, glow, flare, highlight contamination, and local detail degradation coexist around strong artificial light sources.

Extensive experiments demonstrate that the proposed method achieves more faithful restoration of authentic nocturnal appearance than prior nighttime restoration, dehazing, enhancement, and related baselines, both quantitatively and perceptually. It effectively reduces intrusive light pollution, restores natural sky tone and urban structure, and yields visually cleaner nighttime renderings on challenging real cityscapes. We will also release our dataset to facilitate future research on this problem.

In summary, this paper makes the following contributions:
\begin{itemize}
    \item We present a physically grounded light pollution formation model for night cityscapes, explicitly incorporating anisotropic directional light spread and skyline-induced sky glow, two dominant factors overlooked in prior work.
    \item We design a learning framework that combines physically simulated supervision with large-model priors to overcome the absence of paired real data and improve generalization to real nighttime scenes.
    \item We demonstrate that our method restores authentic nighttime appearance more faithfully than prior nighttime restoration, dehazing, enhancement, and related baselines, both quantitatively and perceptually.
\end{itemize}
\begin{figure*}[t]
    \begin{center}
        \includegraphics[width=1.0\textwidth]{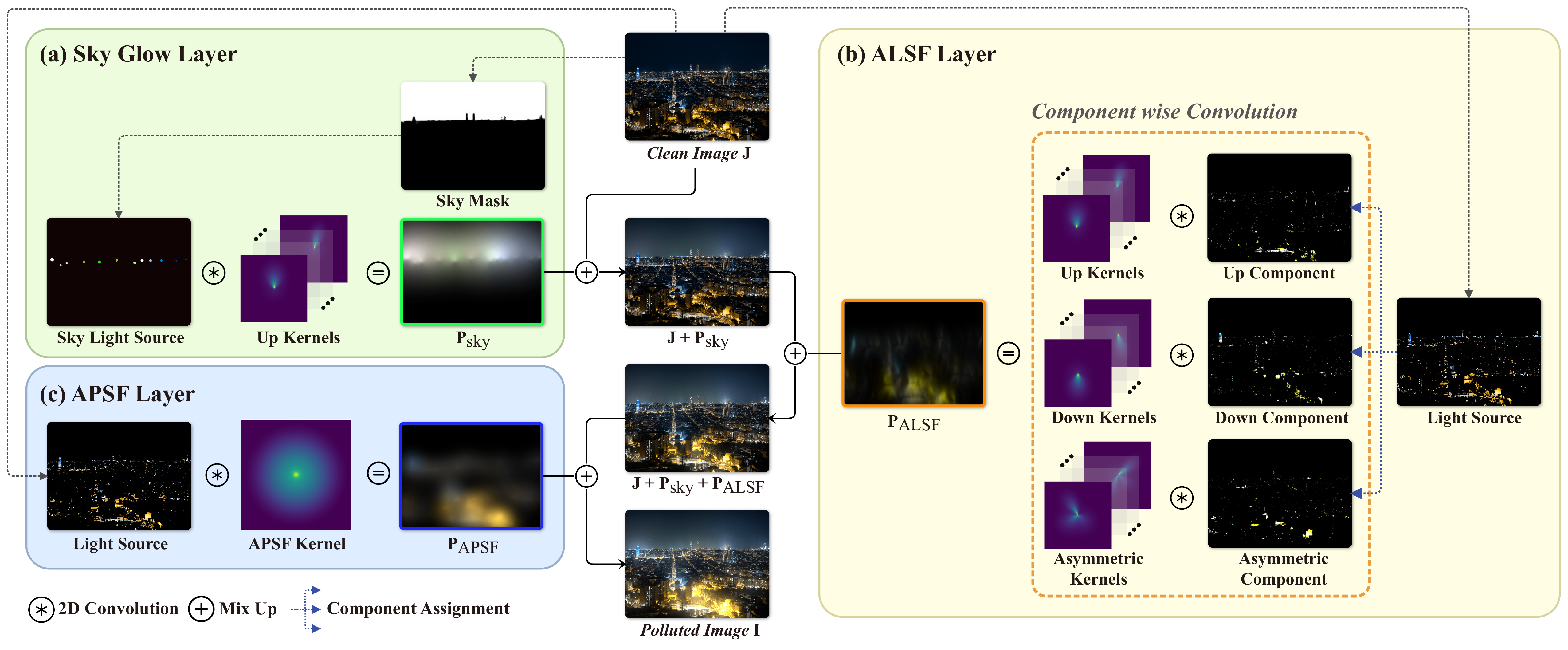}
    \end{center}
    \vspace{-3mm}
    \caption{Overview of the synthetic light-pollution generation pipeline used for dataset construction. From a clean image $J$, we successively add skyline-induced sky glow $P_{\text{sky}}$, component-wise ALSF-based directional scattering $P_{\text{ALSF}}$, and APSF-based isotropic scattering $P_{\text{APSF}}$ to obtain the polluted image $I$.}
    \vspace{-2mm}
    \label{fig:data_pipeline}
\end{figure*}
\section{Related Work}
\label{sec:related}
In this section, we review prior work related to light pollution removal (LPR). The only prior work specifically on LPR is Liu \etal~\cite{liu2021light}, which models pollution as an additive component under isotropic scattering but overlooks anisotropic spread and skyline-induced sky glow, essentially performing perceptual color tuning rather than comprehensive removal. We also review three key related tasks that intersect with LPR: deflare, nighttime dehazing, and low-light image enhancement (LLIE). These tasks address various challenges in improving the quality of nighttime imagery, but often fail to account for the emissive and directional nature of urban light pollution.

\subsection{Deflare}
Deflare methods aim to suppress veiling glare, ghosting, and halo artifacts caused by internal reflections and light scattering within the camera lens. 
Early physics-based approaches~\cite{talvala2007veiling, vitoria2019automatic} modeled flare formation via glare spread functions or geometric priors, followed by spot detection and inpainting for illumination restoration. 
Learning-based methods~\cite{wu2021train, qiao2021light} leveraged synthetic or unpaired data to train neural networks guided by light-source priors for single-image flare removal. 
Recent advances~\cite{ghodesawar2023deflare, dai2023mipi, dai2024mipi, dai2024flare7k++, ma2025self} introduced large-scale datasets, reflection-symmetry priors, and transformer-based architectures to jointly detect and suppress diverse flare patterns. 
DeFlare-Net~\cite{ghodesawar2023deflare} achieves robust flare removal across patterns, while Ma~\etal~\cite{ma2025self} proposed a self-prior-guided spatial-Fourier transformer, and Dai~\etal~\cite{dai2024flare7k++} developed Flare7K++, mixing synthetic and real flares for end-to-end training. 
These methods mainly address local reflections or blooming near light sources but ignore broader pollution like sky glow. Our model integrates local glare with global effects for urban nightscapes.

\subsection{Nighttime Dehazing}
Nighttime dehazing methods aim to mitigate atmospheric scattering and enhance visibility in low-light conditions, where emissive light sources and glow introduce complex degradations like colored haze and noise. Optimization-based approaches~\cite{wang2022variational, ancuti2016night, ancuti2020day, zhang2014nighttime, zhang2017fast, tang2021nighttime, liu2022nighttime} modeled haze formation using atmospheric scattering, gray haze-line priors, maximum reflectance, or variational decomposition to estimate air light and restore illumination, though they often struggle with noise amplification and inaccurate light estimation in complex urban scenes. Learning-based methods~\cite{zhang2020nighttime, yan2020nighttime, kuanar2022multi, jin2023enhancing} employed CNNs, high-low frequency decomposition, or multi-path convolutions trained on synthetic benchmarks to handle domain gaps, but synthetic data limitations hinder generalization to real-world noise and lighting variations. Recent advances~\cite{lin2025nighthaze, cong2024semi, deng2025real} incorporated self-prior learning with severe augmentations, semi-supervised spatial-frequency awareness, and contrastive adversarial training for artifact suppression and robust dehazing. For instance, NightHaze~\cite{lin2025nighthaze} uses degraded clear images for refinement, while Cong~\etal~\cite{cong2024semi} enforces brightness constraints, and Deng~\etal~\cite{deng2025real} bridges synthetic-real gaps. These methods enhance contrast but often under-correct sky glow or leave halos; our LPR distinguishes by modeling anisotropic scattering and glow for aesthetic restoration.

\subsection{Low-light Image Enhancement}
Low-light image enhancement (LLIE) restores visibility, contrast, and color fidelity in underexposed images, often ignoring emissive artifacts from artificial light sources like halos and glow. Traditional methods~\cite{land1977retinex, fu2016weighted, guo2016lime, guo2023low, jiang2024revisiting} used Retinex theory to decompose reflectance and illumination, variational models for map estimation, or illumination-aware adjustments to brighten dark regions, but struggled with noise and over-enhancement in complex scenes. Learning-based approaches advanced the field: supervised methods~\cite{lore2017llnet, wang2023low, wu2023learning, xu2022snr, xu2023low, zamir2020learning, cai2023retinexformer, hai2023advanced, wu2022uretinex, zhang2021beyond, wei2018deep, zhang2019kindling} applied CNNs, autoencoders, and Retinex-inspired networks to learn mappings from paired data for robust restoration; unsupervised and zero-shot techniques~\cite{fu2023learning, guo2020zero, jiang2021enlightengan, liu2021retinex, ma2022toward, yang2023implicit, lee2023temporally, jin2022unsupervised} employed curve estimation, GANs, neural architecture search, or decomposition losses without paired supervision for better generalization; semi-supervised methods~\cite{yang2020fidelity} combined labeled and unlabeled data to balance stability and adaptability. Recent works~\cite{jiang2024lightendiffusion, ma2025incorporating, jiang2023low, zhou2024glare, feijoo2025darkir, yan2025hvi} integrate diffusion models with latent Retinex guidance, Fourier transformations, or wavelet-based architectures for improved quality. While these methods boost visibility effectively, most ignore light pollution, except Jin \etal~\cite{jin2022unsupervised}, who introduced light effects suppression (LES) in LLIE via layer decomposition, guided by tailored losses to suppress artifacts without paired data. However, such LES approaches ignore broader emissive, directional, and skyline-induced urban light pollution, leading to incomplete removal in complex cityscapes. In contrast, the LPR task models these physical factors explicitly, enabling more faithful restoration of authentic nocturnal imagery.
\begin{figure*}[t]
    \begin{center}
        \includegraphics[width=1.0\textwidth]{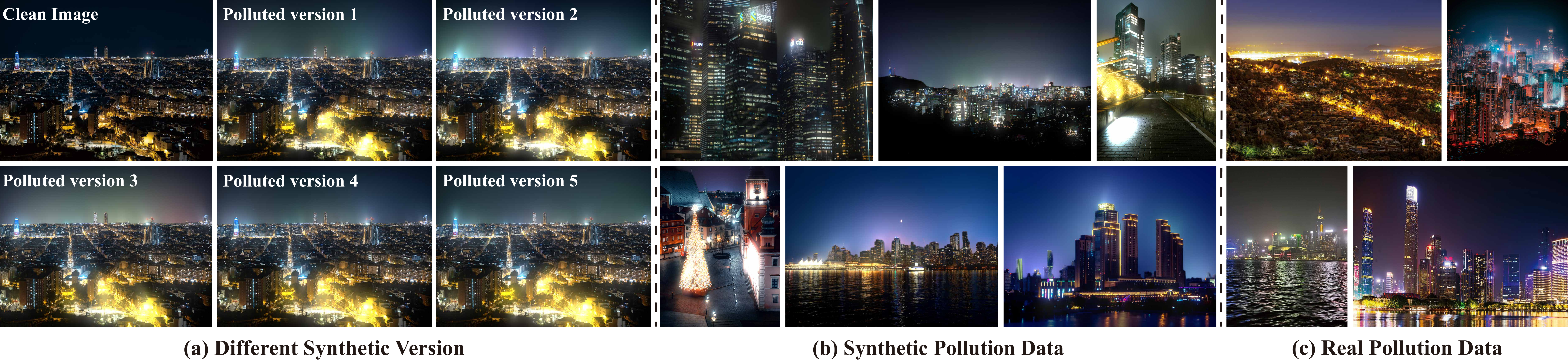}
    \end{center}
    \vspace{-4mm}
    \caption{Examples of our Dataset. (a) Variations of one clean image with simulated light pollution under randomized APSF and ALSF parameters. (b) Additional synthetic polluted samples. (c) Real severely light-polluted nighttime images in the test set.}
    \vspace{-3mm}
    \label{fig:dataset}
\end{figure*}
\section{Method}
\label{sec:method}
Our method tackles the novel LPR task through a physically grounded formation model and a fine-tuning strategy for a large generative model. We model image formation with anisotropic light spread and skyline-induced sky glow (Sec. 3.1), build a physically accurate synthetic dataset (Sec. 3.2), and fine-tune using synthetic supervision with large-model priors for real-scene generalization (Sec. 3.3).

\subsection{Light Pollution Formation Model}
To faithfully capture the radiative and propagative nature of urban light pollution, we propose a composite image formation model that extends the Atmospheric Point Spread Function (APSF)~\cite{narasimhan2003shedding} to account for directional light emitted by the sources. A polluted nighttime image $\mathbf{I}$ is modeled as the sum of the clean nighttime image $\mathbf{J}$ (desired output) and the light pollution components:
\begin{equation}
\mathbf{I}(x) = \mathbf{J}(x) + \mathbf{P}_{\text{sky}}(x) + \mathbf{P}_{\text{APSF}}(x) + \mathbf{P}_{\text{ALSF}}(x)
\label{eq:synthesis_I}
\end{equation}
where $\mathbf{P}_{\text{sky}}$ denotes the diffuse sky glow pollution caused by invisible ground lighting behind skylines scattering into the atmosphere, $\mathbf{P}_{\text{APSF}}$ accounts for the isotropic scattered light from visible sources directed towards the camera, and $\mathbf{P}_{\text{ALSF}}$ represents the anisotropic directional light emitted by visible sources in specific orientations.

\noindent\textbf{Anisotropic Light Spread Function.} 
In real nighttime photography, directional light pollution is common, influenced by factors such as lamp shade constraints, light source emission directions, building occlusions, light source reflections, and dirt or dust on the lens.

To model this anisotropic light spread, we develops an Anisotropic Light Spread Function (ALSF) kernel, building upon the isotropic APSF kernel from Narasimhan \etal~\cite{narasimhan2003shedding}. We start by generating an isotropic APSF kernel based on physical principles, parameterized by optical thickness $T$, scattering coefficient $q$, and kernel size $s$. This kernel is then transformed into an anisotropic form through a warping process designed to introduce directionality. The warping is achieved by creating a displacement field, facilitated by converting from Cartesian to polar coordinates to efficiently model the multi-scattering effects of light starting from a point source.

We design this displacement field to produce comet-like directional tails by modeling one or more shaped light beams emanating from the center. Each shaped light beam is defined by its directional angle $\alpha$, spread angle $\sigma$, and intensity $A$. A global decay factor $\kappa$ is applied to modulate the overall effect with distance. 

For each pixel with offsets $dx$ and $dy$ from the kernel center, we calculate the pixel's polar angle $\theta = \atanTwo(-dy, dx) \cdot 180 / \pi$, normalized to [0, 360). The angular deviation from each beam's angle is then computed as $\delta = \min(|\theta - \alpha|, 360 - |\theta - \alpha|)$, accounting for angular periodicity. The angular contribution from each beam is modeled with a Gaussian function for smooth angular falloff:

\begin{equation}
G = A \cdot \exp\left(-\frac{\delta^2}{\sigma^2}\right)
\label{eq:gaussian_weight}
\end{equation}

We define the scalar warp field $\Phi$ at each pixel as the sum of angular contributions across all beams:

\begin{equation}
\Phi = \sum_{i} G_i
\label{eq:warp_field}
\end{equation}

To modulate with distance, we apply the global decay term based on the radial distance $r = \sqrt{dx^2 + dy^2}$ from the kernel center:

\begin{equation}
D = \exp\left(-\kappa \cdot \frac{r}{w/2}\right)
\label{eq:decay_term}
\end{equation}
where $w/2$ normalizes the distance to the image half-width, ensuring gradual attenuation toward the edges. The displacement field is then constructed as follows: 
\begin{equation}
S' = (1 + \Phi) \cdot D,\quad r' = r / S', \quad\Delta r = r' - r 
\label{eq:radial_displacement}
\end{equation}
first, compute the decayed composite scale factor $S'$, which combines the base scale with the decayed warp field; then, derive the warped source distance $r'$, representing the adjusted radial position after applying the scale; and next, obtain the radial displacement $\Delta r$ as the difference.
We then project the radial displacement back to Cartesian coordinates for the displacement field:
\begin{equation}
\Delta x = \Delta r \cdot \cos(\theta_{\text{rad}}), \quad \Delta y = -\Delta r \cdot \sin(\theta_{\text{rad}})
\label{eq:cartesian_displacement}
\end{equation}
with $\theta_{\text{rad}} = \theta \cdot \pi / 180$, forming the final displacement field $(\Delta x, \Delta y)$ at each pixel.The final ALSF kernel is defined as follows:
\begin{equation}
\text{ALSF}(u,v) = \text{APSF}(u + \Delta x, v + \Delta y) / \text{det}_J(u,v)
\label{eq:ALSF}
\end{equation}
where $\text{APSF}(u,v)$ is the original isotropic APSF kernel, and the Jacobian determinant $\text{det}_J$ compensates for the deformation caused by the displacement field, ensuring energy conservation and maintaining accurate light intensity and distribution across the transformed kernel.

These parameters are varied to simulate the diverse lighting conditions encountered in urban environments, enabling the model to capture the complex and directional nature of light pollution.
\begin{figure}
    \begin{center}
        \includegraphics[width=0.95\linewidth]{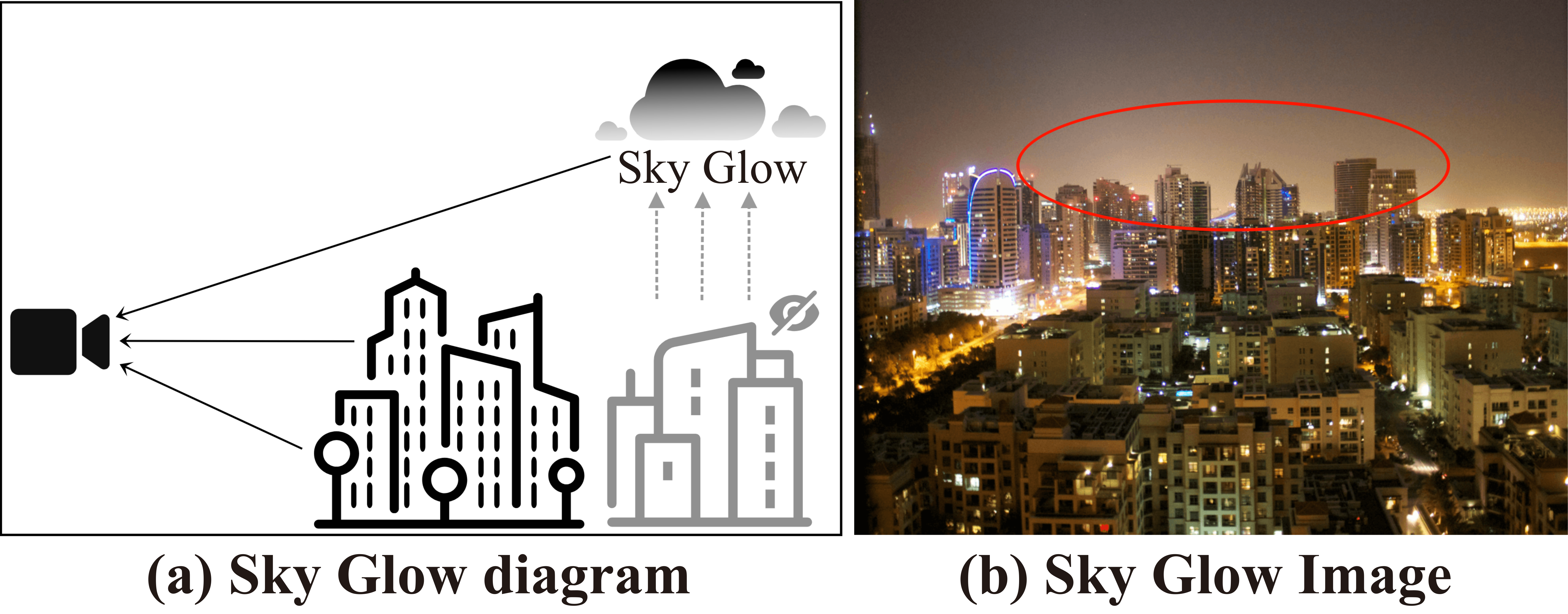}
    \end{center}
    \vspace{-4mm}
    \caption{Illustration of skyline-induced sky glow, where hidden ground lights behind the skyline create a visible glow region in the sky.}
    \label{fig:sky_glow}
    \vspace{-4mm}
\end{figure}
\begin{figure*}[t]
    \begin{center}
        \includegraphics[width=0.95\textwidth]{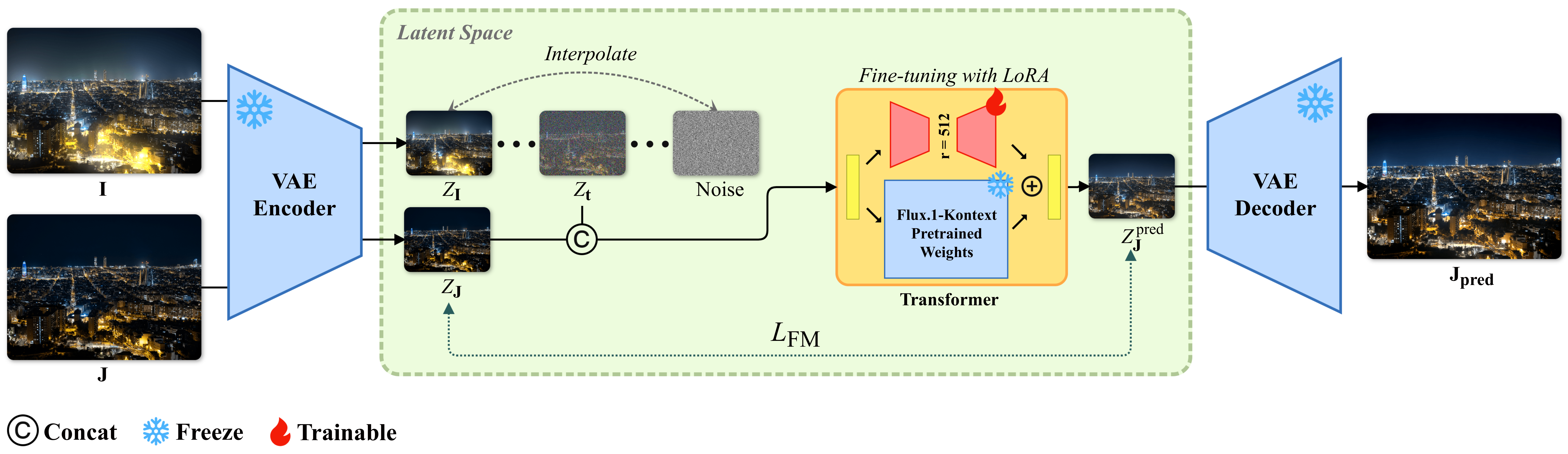}
    \end{center}
    \vspace{-4mm}
    \caption{Overview of our pipeline. The polluted image $I$ and clean target $J$ are encoded into latents; the noisy clean latent is concatenated with the polluted latent and fed into a Q-LoRA fine-tuned Flux.1-Kontext model to predict the clean latent.}
    \vspace{-4mm}
    \label{fig:train_pipeline}
\end{figure*}

\noindent\textbf{Skyline-Induced Sky Glow.}
A critical yet overlooked aspect of light pollution is sky glow from ground lights behind skylines. As shown in Fig.~\ref{fig:sky_glow}, these invisible sources diffusely elevate local sky luminance. Typically distant, they approximate large, upward-facing area lights projecting into the sky, creating localized glow. We simulate this using an upward ALSF kernel for diffuse scattering from distant sources. We first extracts the skyline via a sky-segmentation algorithm, then places varied area lights sources randomly 0 to 15 pixels below it, ensuring no overlap and excluding visible areas to keep them hidden. These light sources are varied in size, shape, color, and intensity. In line with the typical color characteristics of urban lighting, we use five saturated colors with 20\% hue variation for diversity, enabling the synthetic dataset to reflect real-world nighttime variability. This simulation provides realistic urban light pollution representation, addressing a key gap in prior nighttime restoration data.

\subsection{Dataset Construction}
Capturing nighttime photographs free of light pollution is inherently challenging, requiring optimal weather, vantage points, professional equipment, skilled photographers, and extensive post-processing. As a result, no specialized dataset currently exists for LPR tasks. To fill this gap, we construct the first light-pollution-removal dataset by curating 521 high-quality nighttime images with well-controlled illumination and minimal pollution. This set includes about 200 astronomical landscape photos and over 300 urban nighttime scenes. The astronomical images, though not urban, are intentionally included because they are typically captured in remote, pollution-free environments, providing clean priors that are otherwise unattainable in urban photography.

As illustrated in Fig.~\ref{fig:data_pipeline}, our dataset construction begins with preprocessing the curated images. We first segment and mask sky regions using Google’s Sky Optimization algorithm \cite{liba2020sky}, with manual refinement to accurately remove sky-based luminous objects. Ground light sources are then extracted using the method of Levin \etal \cite{levin2007closed}. To synthesize realistic light pollution, we successively overlay three components onto each clean image: a sky-glow layer $\mathbf{P}{\text{sky}}$ capturing diffuse pollution from distant sources, an ALSF layer $\mathbf{P}_{\text{ALSF}}$ modeling directional scattering from visible lights, and an APSF layer $\mathbf{P}_{\text{APSF}}$ simulating uniform forward scattering toward the camera. During synthesis, we randomly sample parameters for APSF (optical thickness $T$, scattering coefficient $q$, kernel size $s$) and ALSF (direction $\alpha$, spread $\sigma$, decay $\kappa$) to ensure diverse, realistic pollution variations.

When simulating $\mathbf{P}_{\text{ALSF}}$, we utilize three primary ALSF kernels, upward direction, downward direction, and multi asymmetric direction, to replicate common urban light spread patterns. To assign these kernels to visible light sources, we segment the light sources into connected components, treating each as a distinct emitter and assigning a unique ALSF kernel. 

For each clean nighttime image $\mathbf{J}$, we synthesize 5 different polluted versions, resulting in approximately 2,500 paired images in total. As illustrated in Fig.~\ref{fig:dataset}(a), each group consists of one clean image and its corresponding polluted variants. Fig.~\ref{fig:dataset}(b) shows additional examples of synthetic pollution data. The dataset is split by group in a 9:1 ratio for training and validation to avoid overlap between variants of the same source image. To assess generalization in real-world scenarios, we additionally collect 100 severely light-polluted urban nighttime images from the web as a test set, with examples depicted in Fig.~\ref{fig:dataset}(c). The dataset will be made publicly available for future research.

\subsection{Model and Training Strategy}
We fine-tune the pre-trained Flux.1-Kontext~\cite{labs2025flux1kontextflowmatching} model to improve robustness to diverse light pollution effects and exploit its generative ability to recover details in overexposed regions. The overall training pipeline is shown in Fig.~\ref{fig:train_pipeline}.
For efficient adaptation, we adopt Q-LoRA, quantizing the base model to 4 bits with a LoRA rank of 512, resulting in about 600 million trainable parameters. During training, a fixed prompt, \textit{“Remove the light glow around light source and make the image look beautiful.”}, is used. The light-polluted image $\mathbf{I}$ and clean target $\mathbf{J}$ are encoded by a VAE into latent codes $Z_{\mathbf{I}}$ and $Z_{\mathbf{J}}$. Following Flow Matching, noise is added to $Z_{\mathbf{J}}$, and the noisy latent is concatenated with $Z_{\mathbf{I}}$ as model input. The network predicts $Z_{\mathbf{J}}^{\text{pred}}$ and is optimized with the Flow Matching loss $L_{FM}$.
We train the model with a batch size of 1 using 256$\times$256 random crops and horizontal flips. The VAE is frozen throughout training. Optimization is performed with Adam~\cite{adam2014method} for 25,000 steps at a learning rate of $10^{-4}$, enabling effective adaptation to synthetic data and good generalization to real nighttime scenes.
\section{Experiments}
\label{sec:exps}
\begin{table}[t]
  \centering
  \caption{Objective metrics comparisons of ours and other methods. The best two scores are shown in \textbf{Bold} and \underline{underline}. \textit{"LES" means "Light Effects Suppression" and "LPR" means "Light Pollution Removal".}}
  \vspace{-2mm}
  \begin{tabular}{l | c | c c }
    \toprule
    \textbf{Method} & \textbf{Task} & \textbf{PSNR} & \textbf{SSIM} \\
    \midrule
    Jin's \textcolor{gray}{(ACM MM 23)}~\cite{jin2023enhancing} & Dehaze & \underline{20.7015} & 0.7757 \\
    Cong's \textcolor{gray}{(CVPR 24)}~\cite{cong2024semi} & Dehaze & 18.1638 & 0.5701 \\
    Lin's \textcolor{gray}{(AAAI 25)}~\cite{lin2025nighthaze} & Dehaze & 16.6349 & 0.5942\\
    Jin's \textcolor{gray}{(ECCV 22)}~\cite{jin2022unsupervised} & LES & 19.3228 & 0.8020 \\
    Dille's \textcolor{gray}{(ECCV 24)}~\cite{dille2024intrinsic} & HDR & 20.1549 & 0.7783 \\
    Dai's \textcolor{gray}{(TPMAI 24)}~\cite{dai2024flare7k++} & Deflare & 19.3228 & \textbf{0.8713} \\
    \midrule
    Liu's \textcolor{gray}{(2021)}~\cite{liu2021light} & LPR & 20.6095 & 0.8314 \\
    \bf{Ours} & LPR & \textbf{26.1728} & \underline{0.8517} \\
    \bottomrule
    \end{tabular}
    \label{tab:Objective metrics}
    \vspace{-4mm}
\end{table}
\begin{figure*}
    \begin{center}
    \includegraphics[width=0.95\textwidth]{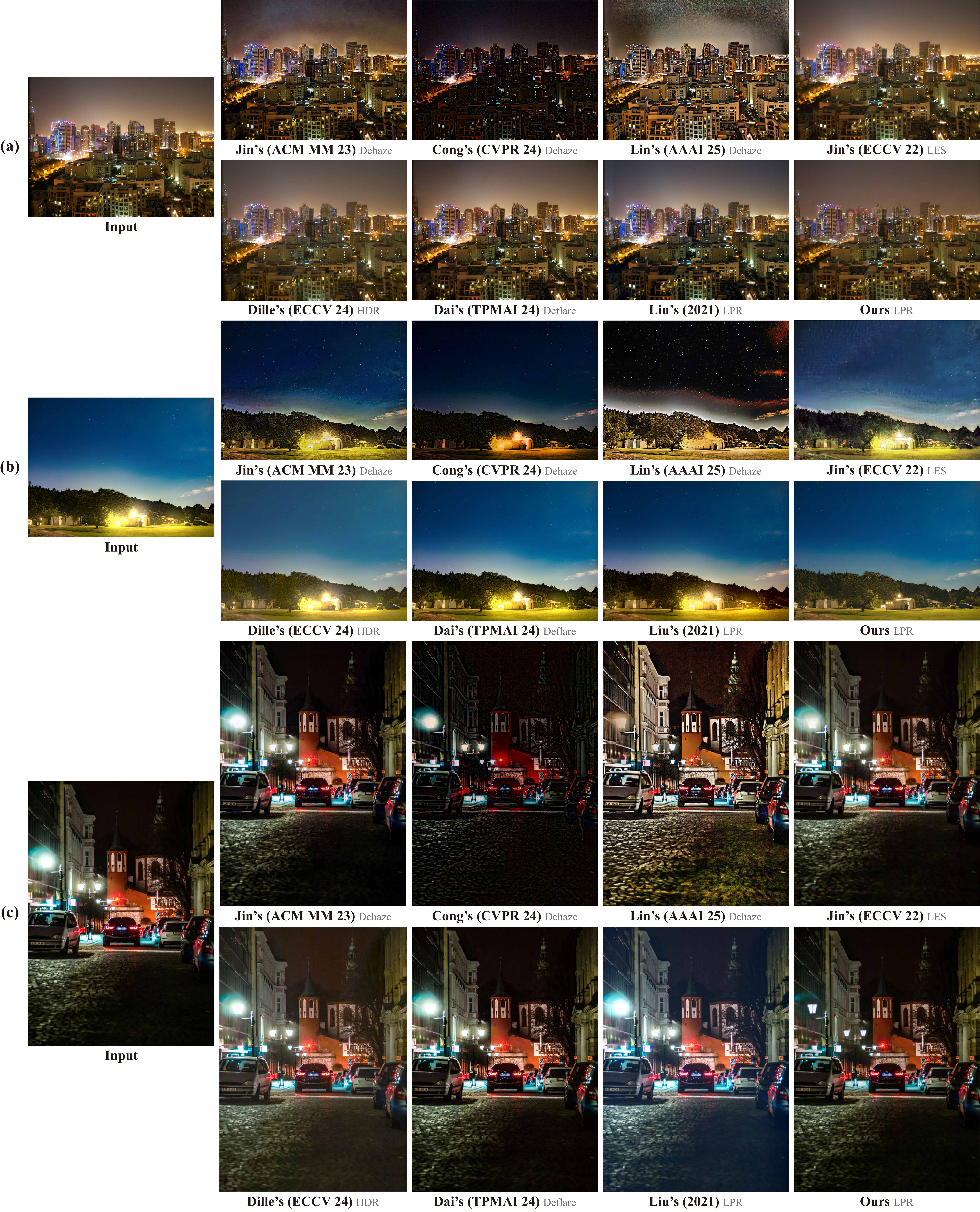}
    \end{center}
    \vspace{-3mm}
    \caption{Qualitative comparison of different methods. Specific tasks are marked in \textcolor{gray}{\scriptsize small grey text}, \textit{"LES" means "Light Effects Suppression" and "LPR" means "Light Pollution Removal".}
    }
    \label{fig:exp_1}
\end{figure*}

\subsection{Comparison with Existing Methods}
To the best of our knowledge, Liu et al.~\cite{liu2021light} remains the only prior work directly addressing LPR. To better contextualize our method, we also compare with recent nighttime enhancement approaches, including nighttime dehazing~\cite{jin2023enhancing, cong2024semi, lin2025nighthaze}, light-effect suppression~\cite{jin2022unsupervised}, and deflaring~\cite{dai2024flare7k++}. Although these methods can reduce light artifacts, they mainly focus on visibility enhancement and often sacrifices color fidelity or aesthetic realism. In contrast, our method is specifically designed to remove light pollution while preserving the natural color palette and nocturnal atmosphere, emphasizing aesthetic consistency rather than visibility-driven enhancement. Since our approach also performs localized exposure correction, we further compare with single-image HDR reconstruction methods such as Dille et al.~\cite{dille2024intrinsic}, which address exposure balancing but are not designed for LPR.

\noindent\textbf{Quantitative comparisons.} We perform a comprehensive assessment using peak signal-to-noise ratio (PSNR) and structural similarity index measure (SSIM) on the validation set of our synthetic dataset. As shown in Table~\ref{tab:Objective metrics}, our method achieves a PSNR of 26.1728 dB and an SSIM of 0.8517, significantly outperforming Liu's~\cite{liu2021light} method. Notably, our PSNR surpasses the second-best (Jin's dehazing at 20.7015 dB) by over 5 dB, indicating superior reconstruction accuracy and reduced artifacts. The high SSIM score, second only to Dai's~\cite{dai2024flare7k++} deflaring method (0.8713), underscores our method's ability to maintain structural integrity while effectively removing light pollution.

\noindent\textbf{Qualitative comparisons.}
Fig.~\ref{fig:exp_1} presents a visual comparison of our method against competing approaches across representative examples.
Nighttime dehazing methods~\cite{jin2023enhancing, cong2024semi, lin2025nighthaze} enhance visibility and sharpness but often distort natural colors and aesthetics. Jin’s~\cite{jin2023enhancing} and Lin’s~\cite{lin2025nighthaze} methods boost contrast and reveal dark details in examples (a) and (b), yet exacerbate sky glow and alter sky colors. In examples (b) and (c), Lin’s~\cite{lin2025nighthaze} suppresses light spill somewhat but changes light source colors, while Jin’s~\cite{jin2023enhancing} preserves colors with limited spill reduction. Cong’s~\cite{cong2024semi} lowers brightness and sharpens details, effectively mitigating sky glow in (a) and (b), but poorly handles source spill.
Jin’s~\cite{jin2022unsupervised} light-effects suppression reduces glow around light sources, as in (b) where it sharpens lamp-under structures, but degrades sky smoothness, color fidelity, and introduces artifacts.
Dille’s~\cite{dille2024intrinsic} HDR method expands dynamic range and clarifies dark details (e.g., foliage in (b), distant mountains/sky in (c)), yet ignores light pollution and recovers bright details poorly.
Dai’s~\cite{dai2024flare7k++} deflaring shows limited LPR efficacy: sky glow persists in (a), with minor glow reduction near sources in (b) and (c), but broader spill and overexposure remain.
Liu’s~\cite{liu2021light} LPR applies physics-based color tuning without removing pollution, leaving source scattering and sky glow. In (a), it blues the upper sky for a less polluted appearance; changes are negligible in clean-sky (b); and it causes blue casts in (c).

Compared to above methods, our method suppresses sky glow and source scattering while preserving colors and structures: seamless sky-glow removal and toned streetlights in (a); cleared source glow and restored architecture in (b); confined spill revealing lamp structure in (c). Unlike others, it avoids color shifts, artifacts, and distortions, balancing LPR with aesthetics.
\begin{figure}
    \begin{center}
        \includegraphics[width=0.95\linewidth]{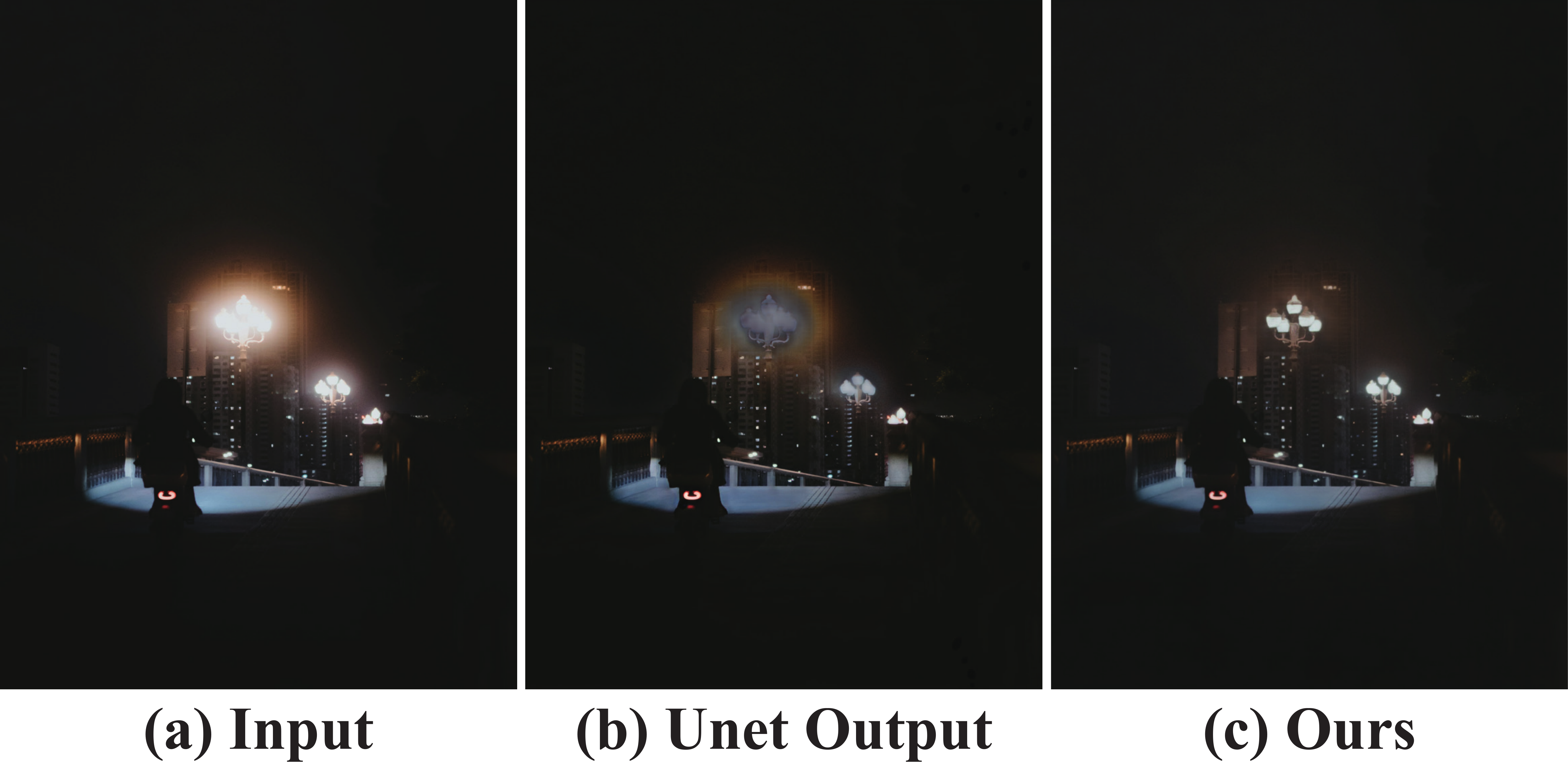}
    \end{center}
    \vspace{-2mm}
    \caption{Backbone ablation showing better overexposed detail recovery with the generative backbone than with UNet.}
    \vspace{-4mm}
    \label{fig:unet_abl}
\end{figure}
\begin{table}[t]
  \centering
  \caption{Ablation studies of ALSF kernel and Sky Glow. The best two scores are shown in \textbf{Bold} and \underline{underline}. }
  \vspace{-2mm}
    \begin{tabular}{c | c c | c c }
        \toprule
        \textbf{Exp.} & $\mathbf{P}_{\text{ALSF}}$ & $\mathbf{P}_{\text{sky}}$ & \textbf{PSNR} & \textbf{SSIM} \\
        \midrule
        (a) & \ding{55} & \ding{55} & 22.0269 & 0.8177 \\
        (b) & \ding{51} & \ding{55} & 23.1589 & 0.8116 \\
        (c) & \ding{55} & \ding{51} & \underline{23.6253} & \underline{0.8341} \\
        (d) & \ding{51} & \ding{51} &\textbf{26.1728} & \textbf{0.8517}\\
        \bottomrule
    \end{tabular}
    \label{tab:Ablation}
    \vspace{-4mm}
\end{table}
\begin{table}
  \centering
  \caption{Average Rankings of Methods. The best average ranks are shown in \textbf{Bold}.}
  \vspace{-2mm}
    \begin{tabular}{ c | c | c }
        \toprule
        \textbf{No.} & \textbf{Method} & \textbf{Average Rank} \\
        \midrule
        1) & Jin's \textcolor{gray}{(ACM MM 23)}~\cite{jin2023enhancing} & 4.67 \\
        2) & Cong's \textcolor{gray}{(CVPR 24)}~\cite{cong2024semi} & 6.38 \\
        3) & Lin's \textcolor{gray}{(AAAI 25)}~\cite{lin2025nighthaze} & 5.11 \\
        4) & Jin's \textcolor{gray}{(ECCV 22)}~\cite{jin2022unsupervised} & 5.18 \\
        5) & Dille's \textcolor{gray}{(ECCV 24)}~\cite{dille2024intrinsic} & 3.92  \\
        6) & Dai's \textcolor{gray}{(TPMAI 24)}~\cite{dai2024flare7k++} & 3.96 \\
        7) & Liu's \textcolor{gray}{(2021)}~\cite{liu2021light} & 4.8 \\
        8) & \textbf{Ours} & \textbf{1.98} \\
        \bottomrule
    \end{tabular}
    \vspace{-4mm}
    \label{table:method_avg_ranking}
\end{table}
\begin{figure*}[t]
    \centering
    \includegraphics[width=0.93\linewidth]{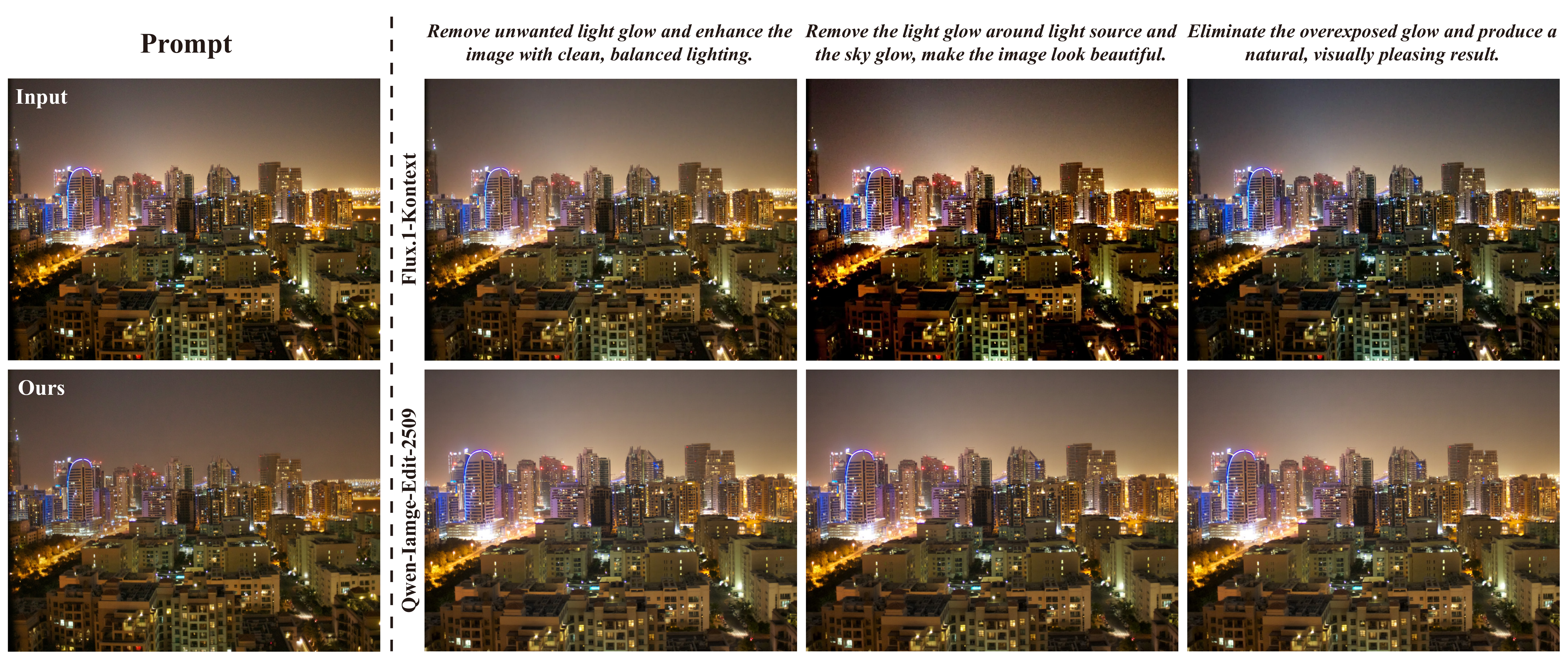}
    \vspace{-4mm}
    \caption{Comparison with instruction-based image editing models, showing that task-specific fine-tuning is needed for severe light pollution removal.}
    \label{fig:compare_mllm}
    \vspace{-4mm}
\end{figure*}
\label{sec:ablation}
\subsection{Ablation Study}
We evaluate the contribution of the two physical terms in our formation model (Eq.~\ref{eq:synthesis_I}) under an identical training setup. In all variants the isotropic component $\mathbf{P}_{\text{APSF}}$ is retained, while the directional term $\mathbf{P}_{\text{ALSF}}$ and the skyline-induced component $\mathbf{P}_{\text{sky}}$ are toggled to match Table~\ref{tab:Ablation}: (a) only $\mathbf{P}_{\text{APSF}}$; (b) $\mathbf{P}_{\text{APSF}}{+}\mathbf{P}_{\text{ALSF}}$; (c) $\mathbf{P}_{\text{APSF}}{+}\mathbf{P}_{\text{sky}}$; and (d) $\mathbf{P}_{\text{APSF}}{+}\mathbf{P}_{\text{ALSF}}{+}\mathbf{P}_{\text{sky}}$. On the same validation split, the baseline (a) reaches 22.0269~dB PSNR / 0.8177 SSIM. Adding $\mathbf{P}_{\text{ALSF}}$ in (b) increases PSNR to 23.1589 with a comparable SSIM, while introducing $\mathbf{P}_{\text{sky}}$ in (c) yields 23.6253~dB / 0.8341, underscoring the importance of modeling skyline-driven background lift. The full configuration (d) attains the best results, \textbf{26.1728~dB} PSNR and \textbf{0.8517} SSIM, corresponding to absolute gains of +4.1459~dB and +0.0340 over (a). 
Furthermore, as shown in Fig.~\ref{fig:unet_abl}, replacing the backbone with UNet fails to recover details in overexposed regions, highlighting the advantage of leveraging a large generative model in handling severe light pollution artifacts. Taken together, these observations indicate that the proposed method provides a stronger physical constraint and effectively suppresses light pollution, yielding more faithful restoration.
\subsection{Compare with Image Editing Models}
We compare our method with two state-of-the-art instruction-based image editing models: FLUX.1-Kontext~\cite{labs2025flux1kontextflowmatching} and Qwen-Image-Edit-2509~\cite{wu2025qwenimagetechnicalreport}. 
Despite testing various prompts to remove light glow and overexposure (three examples in Fig.~\ref{fig:compare_mllm}), both models show limitations.
Qwen-Image-Edit-2509 outputs are nearly identical to inputs, failing to follow instructions.
FLUX.1-Kontext increases contrast and sometimes shifts the sky to cooler tones, mildly reducing orange skyglow, but diffuse dome and halos persist.
Our fine-tuned model eliminates light pollution, restoring a natural sky with sharp, glow-free city lights. This highlights general models' zero-shot struggles and the need for task-specific fine-tuning.

\subsection{User Study}
\begin{figure}[h]
    \centering
    \includegraphics[width=0.95\linewidth]{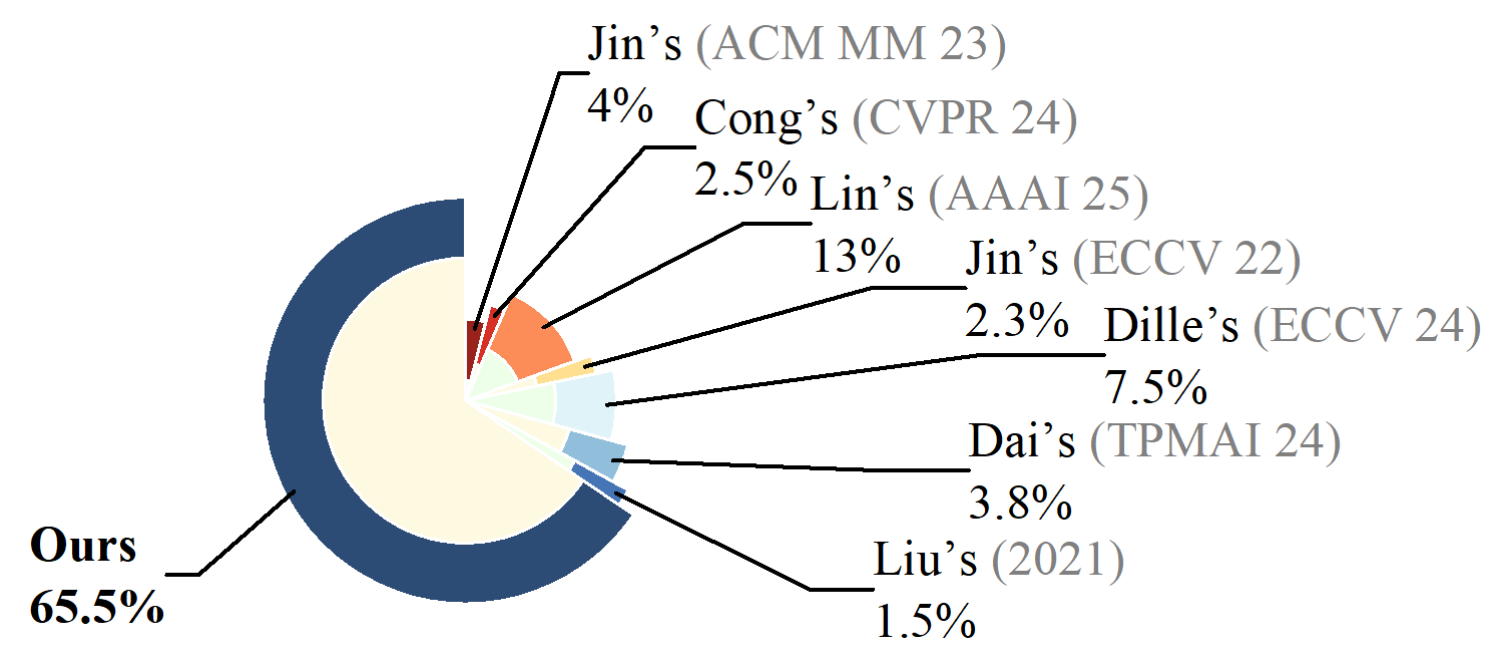}
    \caption{Top-1 ranking rates for different methods.}
    \label{fig:user_study}
    \vspace{-6mm}
\end{figure}

To comprehensively assess the perceptual quality of our LPR results, we conducted a subjective user study on 28 groups of severely polluted nighttime scenes, including 21 real photographs and 7 synthetic validation images. Each group contained one input image and eight restored results produced by our method and seven competing methods. For each participant, 10 groups were randomly sampled, and the order of the eight results in each group was shuffled. Participants were asked to rank the results from 1 (best) to 8 (worst) according to two criteria: (1) effective suppression of light sources and glow without unnatural halos or over-darkening, and (2) overall visual naturalness, including color balance, brightness, and detail clarity.
Eighty participants completed the study, yielding 6,400 ranking entries in total. As shown in Table~\ref{table:method_avg_ranking}, our method achieved the best average rank among all methods. In addition, Fig.~\ref{fig:user_study} indicates that our approach obtained the highest first-place rate for most scenes. These results demonstrate the clear superiority of our method. Detailed statistics are provided in the supplementary materials.

\section{Conclusion}
In this paper, we introduced a novel approach to light pollution removal in nighttime cityscape photographs, addressing the limitations of prior methods by formulating a physically grounded model that explicitly incorporates anisotropic directional light spread via ALSF kernels and skyline-induced sky glow. To overcome the scarcity of paired real data, we constructed a synthetic dataset through physically accurate simulation and coupled it with large-model priors for effective fine-tuning of Flux.1-Kontext using Q-LoRA. Extensive experiments, including quantitative metrics, qualitative comparisons, ablation studies, and a user study, demonstrate that our method outperforms existing nighttime deflare, dehazing, LES, and HDR baselines in LPR task while preserving authentic nocturnal aesthetics.

Our work highlights the importance of modeling overlooked physical factors for realistic nighttime light pollution removal. However, our approach does not include color tuning, so the resulting images may appear less visually appealing in terms of color vibrancy and contrast compared to dedicated HDR methods. Additionally, due to the use of a diffusion model, inference times are not real-time. Future directions include extending the model to handle dynamic scenes, such as videos, and integrating real-time processing for practical applications in nighttime photography, visual arts, and digital image enhancement.

\putbib[arxiv/sample-base]
\end{bibunit}

\clearpage
\begin{figure*}[!t]
\centering
\includegraphics[width=0.96\textwidth]{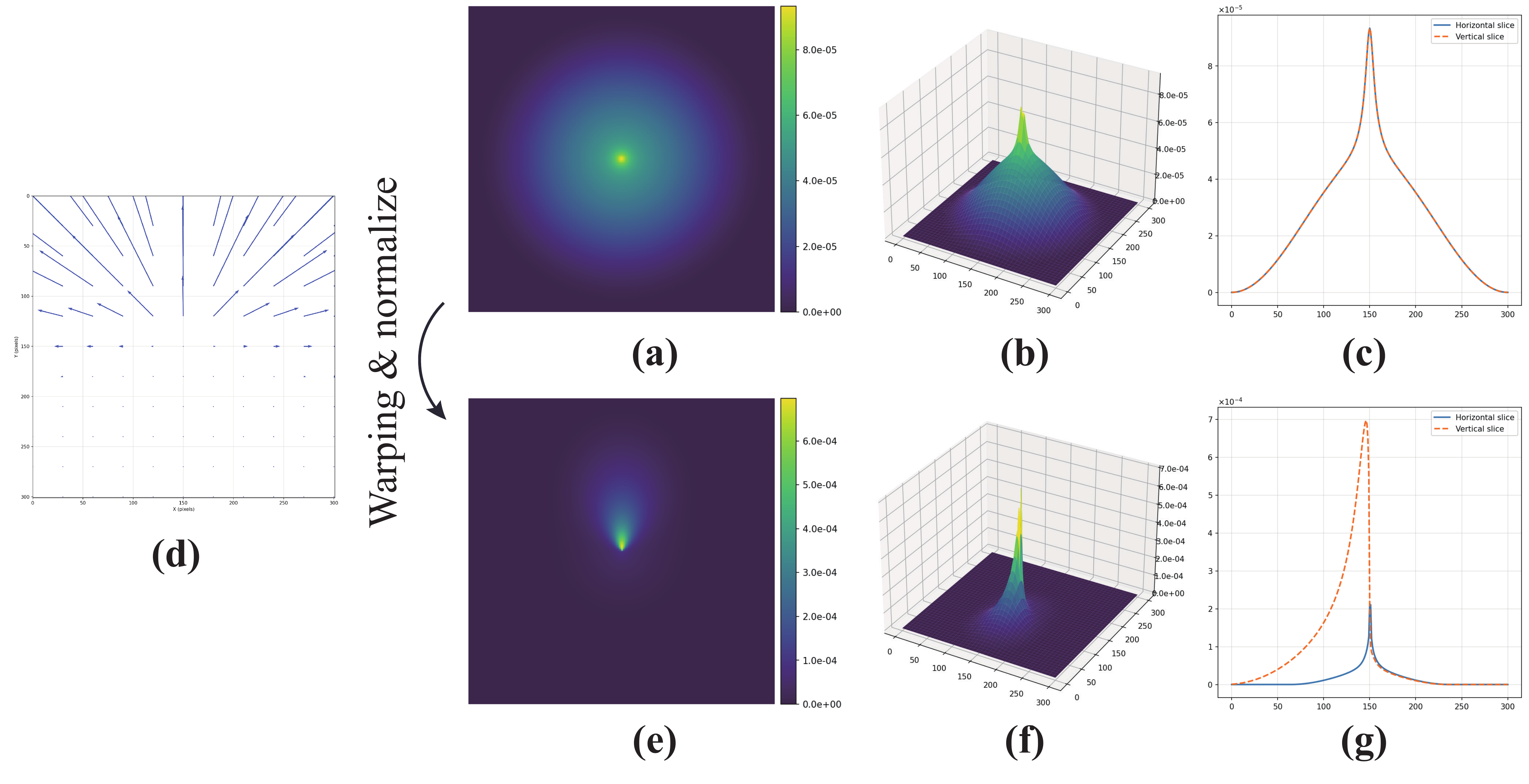}
\caption{Visualization of the ALSF construction process.}
\label{fig:alsf_construction}
\end{figure*}
\appendix
\section*{Supplementary Material}
\begin{bibunit}
In this document, we further elaborate on additional details of our work, including: 1) the design details of ALSF, 2) detailed component assignment in the ALSF layer, 3) a complete analysis of user study and 4) more qualitative comparison results.
\section{Design Details of ALSF}
To offer a deeper understanding of the Anisotropic Light Spread Function (ALSF) introduced in the main paper (Sec.~3.1), we detail its construction by warping an isotropic APSF kernel~\cite{narasimhan2003shedding} with a direction-dependent field to better model anisotropic light propagation in urban night scenes, including intermediate visualizations and chosen hyperparameters.

The process begins with the generation of an isotropic APSF kernel, parameterized by optical thickness $T$, scattering coefficient $q$, and kernel size $s$. This kernel captures uniform multi-scattering effects from a point source. Next, we introduce anisotropy via a warping process that models directional beams emanating from the light source. This is achieved by constructing a displacement field in polar coordinates, as detailed in the main paper. The field incorporates one or more beams defined by directional angle $\alpha$, spread angle $\sigma$, and intensity $A$ (set to 2.0 in our implementation), with a global decay factor $\kappa$ modulating the effect over distance. The scalar warp field $\Phi$ (Eq.~(3) in main paper) aggregates Gaussian contributions from each beam (Eq.~(2)), and the decay term $D$ (Eq.~(4)) ensures gradual attenuation. The resulting radial displacement $\Delta r$ (Eq.~(5)) is projected back to Cartesian coordinates (Eq.~(6)) to form the displacement field $(\Delta x, \Delta y)$.

The final ALSF kernel is obtained by resampling the APSF kernel according to the displacement field and normalizing by the Jacobian determinant $\det J$ to preserve energy (Eq.~(7) in main paper).
Fig.~\ref{fig:alsf_construction} illustrates the entire construction pipeline: The isotropic APSF kernel is visualized in (a) as a 2D heatmap, where brighter regions indicate higher intensity; (b) as a 3D surface plot for a volumetric view; and (c) with cross-sections along the horizontal (blue) and vertical (orange) axes through the kernel center, highlighting the symmetric decay. The warping displacement field is shown in (d) as a 2D vector plot, where arrows indicate the direction and magnitude of displacement at each pixel relative to the kernel center, revealing comet-like tails induced by the directional parameters to simulate constrained light emission (e.g., due to lamp shades or occlusions). After warping and normalization, the resulting ALSF kernel is depicted in (e) as a 2D heatmap showing asymmetric spread; (f) as a 3D surface plot emphasizing directional tails; and (g) with cross-sections along the horizontal (blue) and vertical (orange) axes, illustrating the anisotropy compared to the isotropic case.

In our experiments, parameters are randomly sampled during dataset synthesis to ensure diversity and realism. For the APSF kernel:
\begin{itemize}
\item Optical thickness $T \sim \mathcal{U}(1.1, 1.8)$,
\item Scattering coefficient $q \sim \mathcal{U}(0.2, 0.7)$,
\item A scaling factor $\text{scalor} \sim \mathcal{U}(0.75, 1.5)$, with kernel size $s = \lfloor \text{scalor} \times \max(H, W) \rfloor$, where $H$ and $W$ are the image height and width.
\end{itemize}
For the ALSF warping (using a single beam for primary directions, with extensions to multi-beam in asymmetric cases):
\begin{itemize}
\item Direction angles $\alpha \sim \mathcal{U}(75, 105)$ (in degrees, centered around upward/vertical for common urban lighting),
\item Spread angles $\sigma \sim \mathcal{U}(15, 60)$ (in degrees, to model varying beam widths),
\item Decay factor $\kappa \sim \mathcal{U}(0.5, 1.0)$ (to control radial attenuation strength).
\end{itemize}
These ranges were determined through experiments to replicate observed light pollution patterns in real nighttime cityscapes, such as upward spills from streetlights or asymmetric glows from occluded sources. Multi-beam configurations extend this by sampling multiple $\alpha$ and $\sigma$ values, ensuring comprehensive coverage of directional effects.

\section{Component Assignment in the ALSF Layer}
To generate the anisotropic light pollution component $P_{\text{ALSF}}$ in the dataset construction pipeline (Fig.~2(b) in the main paper), we process the light source map by extracting connected components from it and assigning them to different ALSF kernel types for component-wise convolution. This approach simulates diverse directional light spreads observed in real urban scenes.
The procedure involves the following steps:

\textbf{Extract Connected Components}: The light source map is first binarized using a threshold to create a mask, followed by connected component labeling with 8-connectivity via a $3\times3$ neighborhood. Small components below a minimum area threshold are then filtered out.

\textbf{Assign Kernel Types}: Each identified connected component is randomly assigned to one of the three kernel types through uniform sampling over $\{0, 1, 2\}$, corresponding to upward, downward, and asymmetric kernels, respectively, thereby establishing a mapping from component indices to kernel types.

\textbf{Group by Kernel Type}: For each kernel type, the intensities of the assigned components are aggregated to form grouped light source maps.

\textbf{Component-wise Convolution}: For each non-empty grouped map, frequency-domain convolution is performed with the corresponding kernel after padding to half the kernel size and repeating it across channels, with results accumulated to form the glow layer.

This process produces the $P_{\text{ALSF}}$ layer along with component statistics. The random kernel assignment introduces diversity into the synthetic dataset, improving the model's generalization to varied directional light pollution in real images.

\begin{figure}
    \centering
    \includegraphics[width=1.0\linewidth]{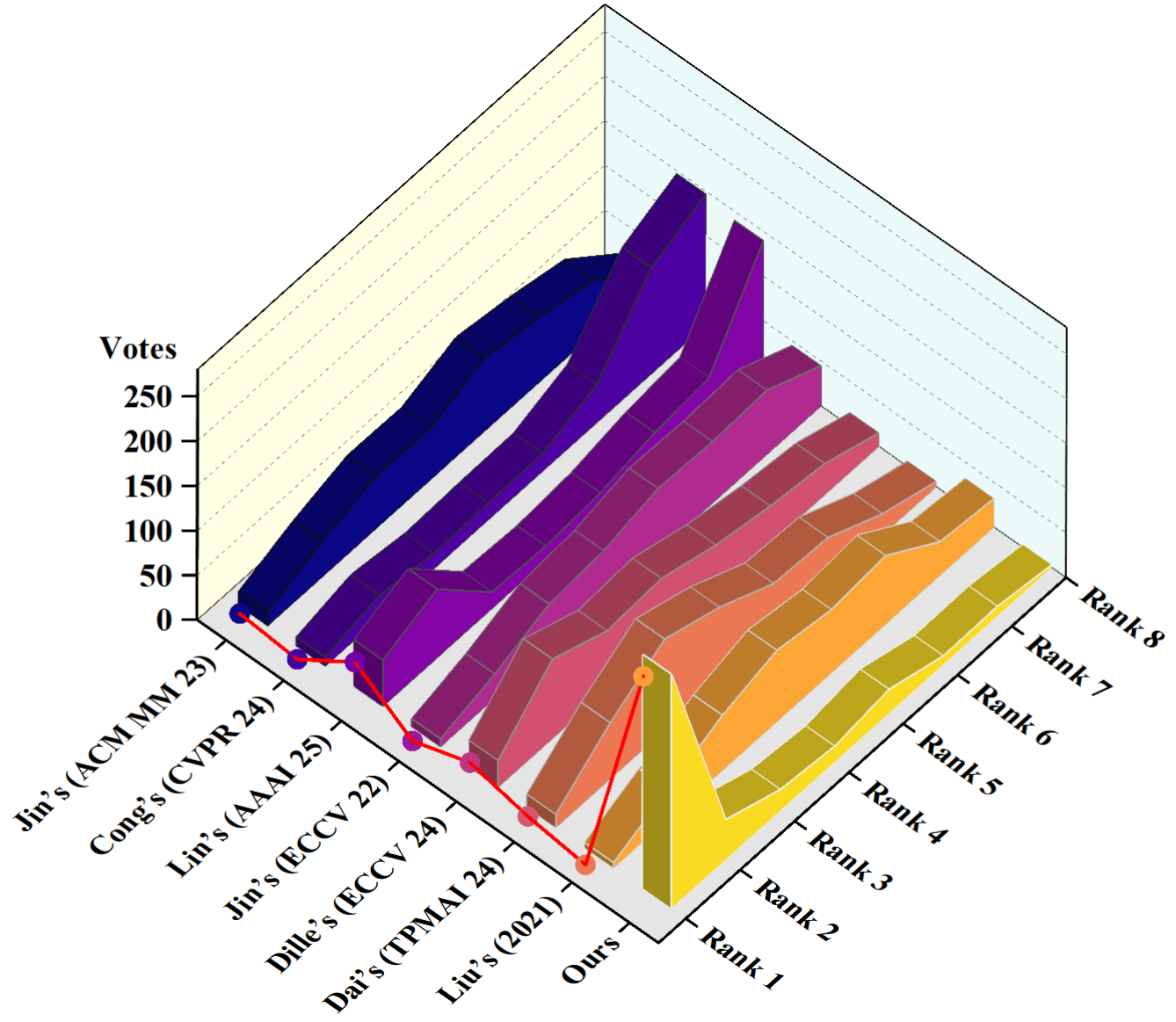}
    \caption{3D bar chart illustrating the distribution of ranks assigned to each method in the user study.}
    \vspace{-2mm}
    \label{fig:user_study_sup}
    \vspace{-4mm}
\end{figure}
\section{Analysis of User Study}
Fig.~\ref{fig:user_study_sup} visualizes the complete voting records from our user study, where 80 participants ranked the restoration results of eight methods, including ours. The data is presented as a 3D bar chart, with methods along one axis, ranks 1 to 8 along another, and vote counts along the third. This allows analysis from two perspectives. Along the method axis, our approach receives the highest number of first-place votes at 524, significantly outperforming others. Along the ranking axis, our method has the greatest probability of being ranked first at 65.5\%, highlighting its consistent preference among users for superior light pollution removal and natural nocturnal restoration.


\section{More Qualitative Comparisons}
\begin{figure*}[h!]
    \begin{center}
        \includegraphics[width=1.0\linewidth]{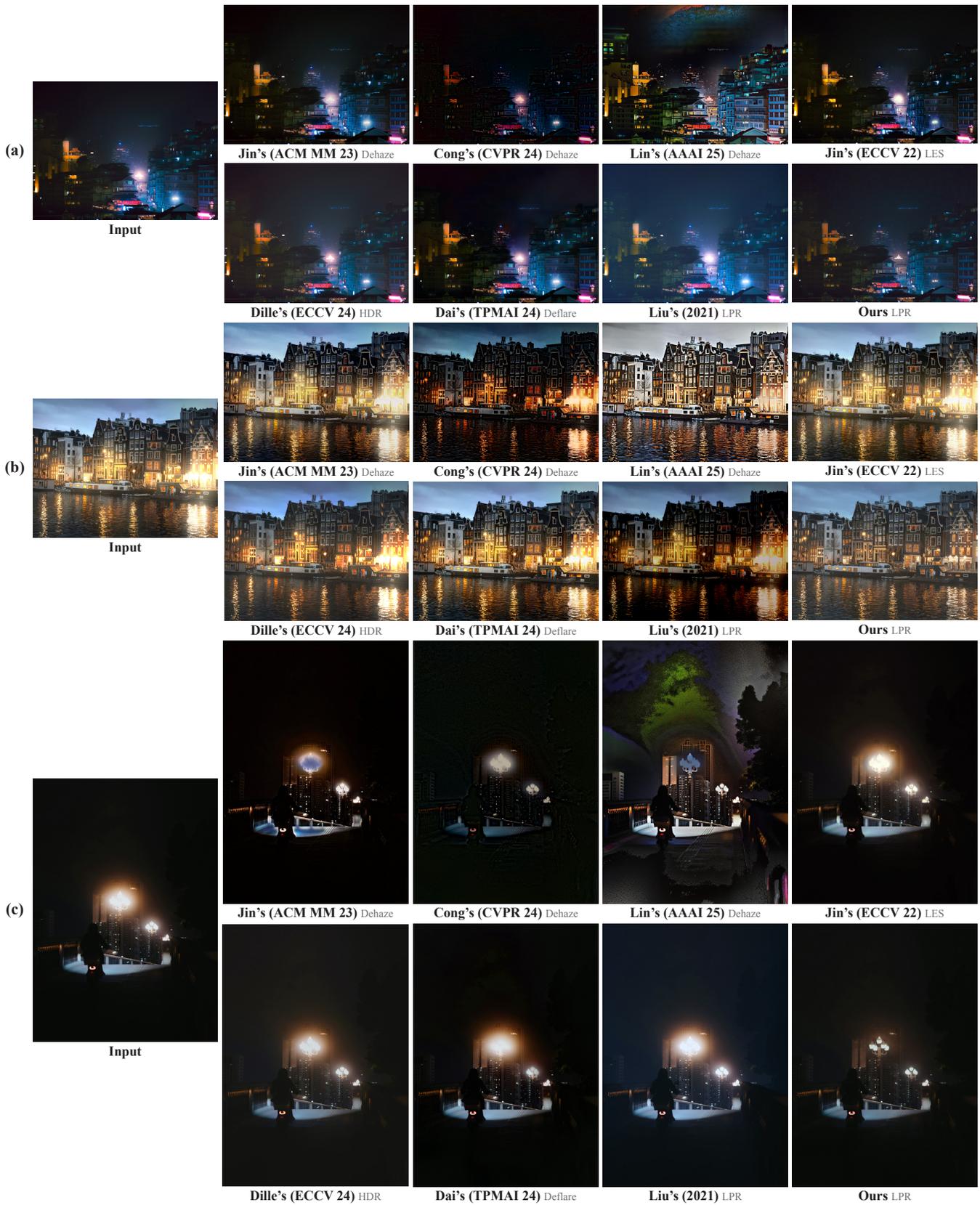}
    \end{center}
    \vspace{-3mm}
    \caption{Qualitative comparison of different methods. Specific tasks are marked in \textcolor{gray}{\scriptsize small grey text}, \textit{"LES" means "Light Effects Suppression" and "LPR" means "Light Pollution Removal".} Zoom in for better view.}
    \label{fig:suppl_1}
\end{figure*}

\begin{figure*}[h!]
    \begin{center}
        \includegraphics[width=1.0\linewidth]{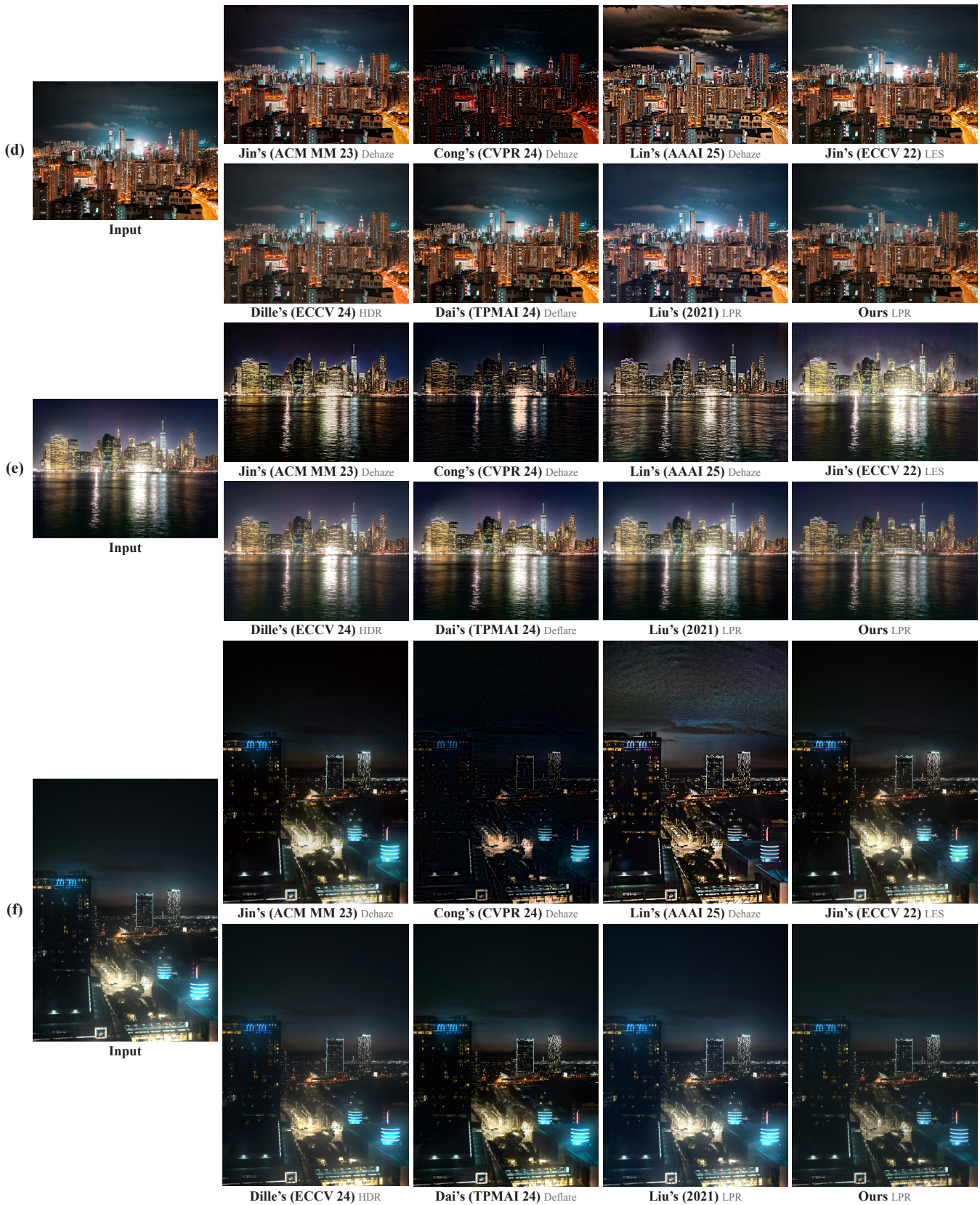}
    \end{center}
    \vspace{-3mm}
    \caption{Qualitative comparison of different methods. Specific tasks are marked in \textcolor{gray}{\scriptsize small grey text}, \textit{"LES" means "Light Effects Suppression" and "LPR" means "Light Pollution Removal".} Zoom in for better view.}
    \label{fig:suppl_2}
\end{figure*}

\begin{figure*}[h!]
    \begin{center}
        \includegraphics[width=1.0\linewidth]{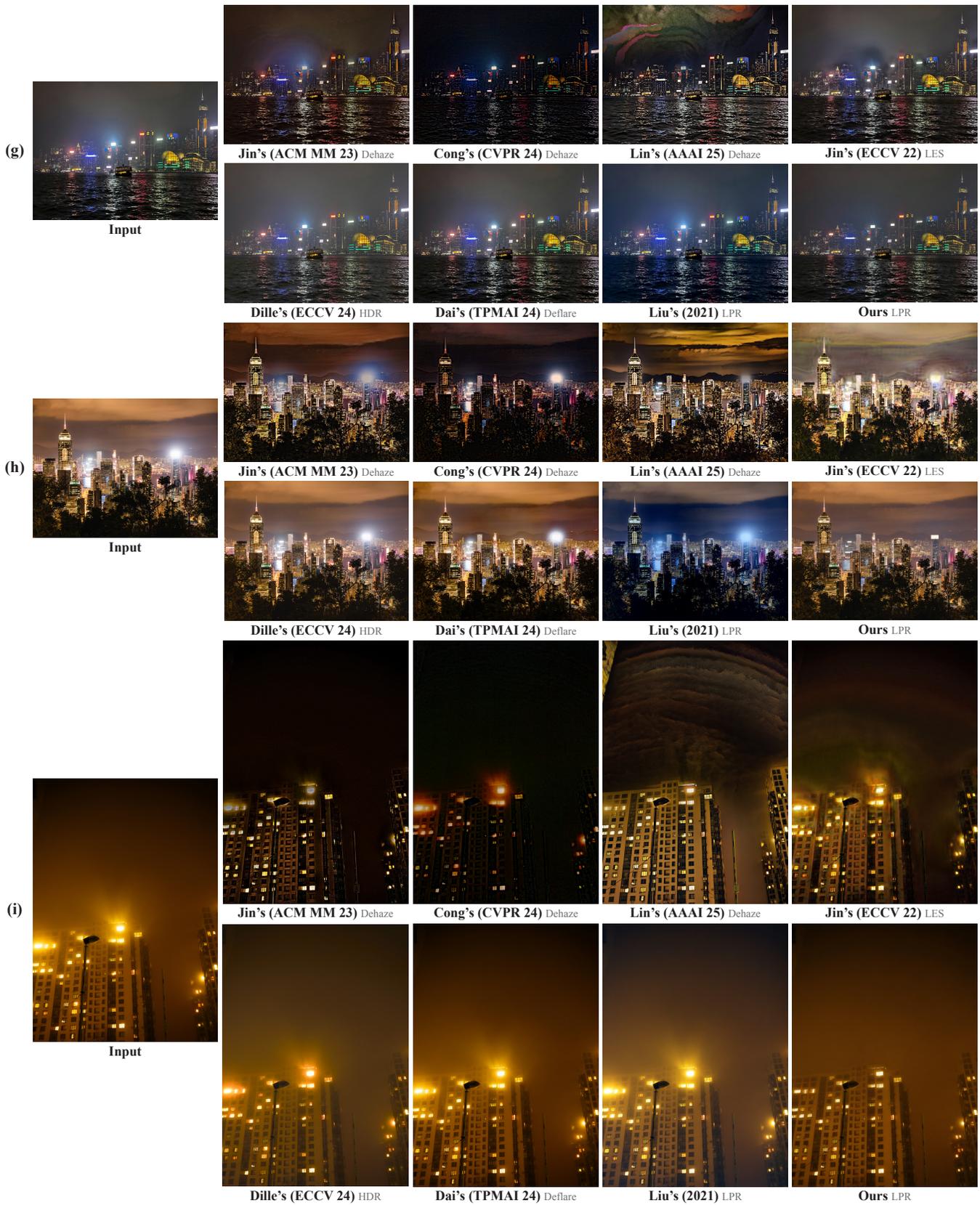}
    \end{center}
    \vspace{-3mm}
    \caption{Qualitative comparison of different methods. Specific tasks are marked in \textcolor{gray}{\scriptsize small grey text}, \textit{"LES" means "Light Effects Suppression" and "LPR" means "Light Pollution Removal".} Zoom in for better view.}
    \label{fig:suppl_3}
\end{figure*}

\begin{figure*}[h!]
    \begin{center}
        \includegraphics[width=1.0\linewidth]{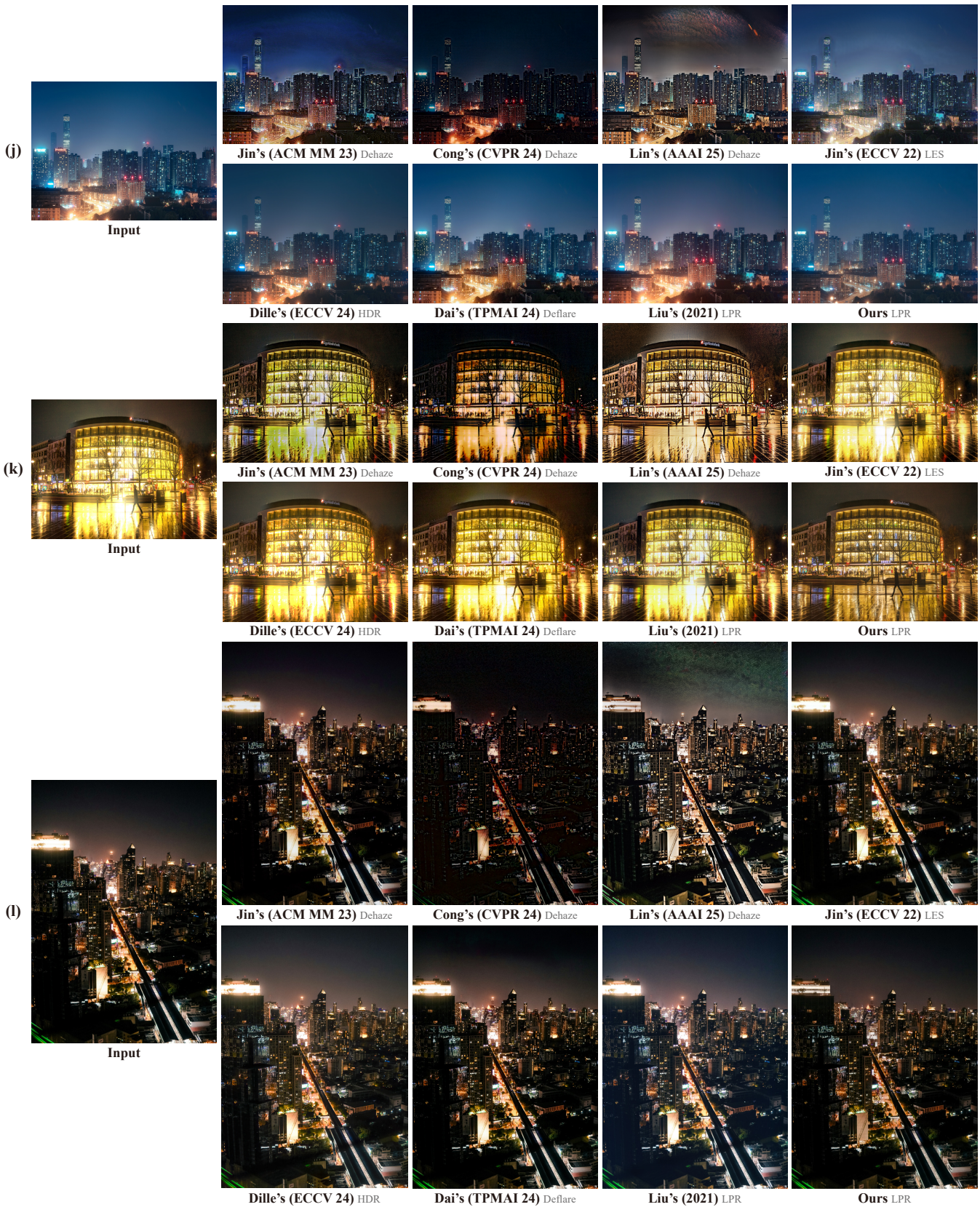}
    \end{center}
    \vspace{-3mm}
    \caption{Qualitative comparison of different methods. Specific tasks are marked in \textcolor{gray}{\scriptsize small grey text}, \textit{"LES" means "Light Effects Suppression" and "LPR" means "Light Pollution Removal".} Zoom in for better view.}
    \label{fig:suppl_4}
\end{figure*}

We provide additional qualitative comparisons in Fig.~\ref{fig:suppl_1}-\ref{fig:suppl_4} with the same competing methods as in the main paper: nighttime dehazing approaches~\cite{jin2023enhancing, cong2024semi, lin2025nighthaze}, light-effects suppression~\cite{jin2022unsupervised}, HDR method~\cite{dille2024intrinsic}, deflaring~\cite{dai2024flare7k++}, and physics-based LPR~\cite{liu2021light}. These extra examples further demonstrate the robustness of our method across diverse nighttime scenes.

\renewcommand{\bibsection}{\section*{References for Supplementary Material}}
\putbib[arxiv/sample-base]
\end{bibunit}


\end{document}